\documentclass[sigconf,screen,authorversion,nonacm]{acmart}

\usepackage{graphicx}
\usepackage{wrapfig}
\usepackage{color}

\begin{document}

\title{Interactive robots as inclusive tools\\to increase diversity in higher education}

\author{Patrick Holthaus}
\orcid{0000-0001-8450-9362}
\affiliation{%
  \institution{University of Hertfordshire}
  \department{School of Physics, Engineering and Computer Science}
  \streetaddress{College Lane}
  \city{Hatfield}
  \country{UK}
  \postcode{AL10 9AB}
}
\email{p.holthaus@herts.ac.uk}

\begin{abstract}
There is a major lack of diversity in engineering, technology, and computing subjects in higher education. The resulting underrepresentation of some population groups contributes largely to gender and ethnicity pay gaps and social disadvantages. We aim to increase the diversity among students in such subjects by investigating the use of interactive robots as a tool that can get prospective students from different backgrounds interested in robotics as their field of study. For that, we will survey existing solutions that have proven to be successful in engaging underrepresented groups with technical subjects in educational settings. Moreover, we examine two recent outreach events at the University of Hertfordshire against inclusivity criteria. Based on that, we suggest specific activities for higher education institutions that follow an inclusive approach using interactive robots to attract prospective students at open days and other outreach events. Our suggestions provide tangible actions that can be easily implemented by higher education institutions to make technical subjects more appealing to everyone and thereby tackle inequalities in student uptake.
\keywords{Equality, Diversity and Inclusion \and Higher Education \and Outreach}
\end{abstract}

\maketitle

\section{Introduction}
Underrepresented groups in Science, Technology Engineering, and Mathematics (STEM) subjects in higher education include women, people from disadvantaged social or ethnically diverse backgrounds, and people with disabilities~\cite{CaSE2014Improving}. According to \cite{HESA2022students}, less than 20 percent of engineering and technology and computing students are female. Such a lack of diversity not only largely contributes to social inequalities, for example, the gender pay gap~\cite{chevalier2004motivation}, but it also constitutes a missed opportunity in terms of economic strength and growth~\cite{allen2014reimagining}. Diversified workplaces on the other hand experience more positive working environments, especially on the individual level, promoting self-esteem and identity expression, which can lead to other positive effects like an increased amount of teamwork~\cite{roberson2019diversity}.

Under-representation at the university is either caused by imbalanced student intake or by disproportional dropout rates. In the United Kingdom (UK), dropout rates in computer science are similar for females (14\%) as for males (15\%)~\cite{smith2001dropping}, which is consistent with the findings of Pappas et al.~\cite{pappas2016investigating} who cannot determine a noteworthy influence of gender on dropout rates in Norway. Still, some countries like Germany observe a 23\% higher dropout rate for females in STEM subjects~\cite{isphording2019gender}. While they suggest some improvement during university studies, they partially explain the higher dropout with lower participation in advanced maths and physics courses during earlier education, i.e. in secondary school settings. Consequently, addressing imbalances at a younger age prior to choosing a study subject seems to provide the most potential for improving diversity at the university.
Equal student intake depends on whether there is an equal interest between the groups in the subject area, which can be heavily influenced by stereotypes about the subject area. \cite{master2016computing} conducted two experiments that found stereotypes of the field can lead to a lowered interest of girls in technical subjects preventing them to enrol in computer science courses. Likewise, \cite{makarova2019gender} find that gender-science stereotypes of maths and science might have an influence on how boys and girls decide to take on a STEM subject. They also show that exhibiting a less pronounced masculine image of science has the potential to increase girls’ interest in a technical subject.

We believe that interactive robots can contribute to increasing diversity in computing subjects by reducing stereotypes about the subject area and by eliciting interest in those prospective students who might have otherwise not considered such a field of study. Interactive robots could be used to attract prospective students for a number of reasons. Firstly, they can draw attention to passing-by people~\cite{saad2019welcoming} and are powerful tools for advertisement~\cite{shiomi2013recommendation}. Secondly, there is evidence that robots can be successfully used in various educational settings~\cite{mubin2013review}. Finally, interactive robots are shaped by humans and their behaviours are designed by humans and, as such, they can be designed to facilitate inclusion, for example, when co-designed with visually impaired and non-impaired children~\cite{neto2021using}. In fact, increasing efforts are undertaken to design robots in an inclusive way~\cite{mohan2015designing}.

To investigate how interactive robots could be used as an inclusive tool to motivate students to pick up robotics research as a subject, we present a literature review in Sect. 2 about papers that present the successful engagement of underrepresented groups using interactive robots in educational settings. In Sect. 3, we present a case study where we analyse outreach activities that use interactive robots at the University of Hertfordshire against inclusivity criteria. We will then summarise and discuss these activities and identify actions that can facilitate engagement with computer science subjects using an interactive robot during outreach events in Sect. 4 and conclude the paper in Sect. 5.

\section{Related Work}
Mubin et al.~\cite{mubin2013review} present a review on the applicability of interactive robots for educational use in schools. They conclude that bringing a robot into the classroom can have stimulating, engaging, and instructive effects. Moreover, interactive robots with their physical presence can provide a hands-on experience that facilitates learning~\cite{belpaeme2018social} and in general, supports a deeper understanding of the subject area~\cite{Beaty2008Supporting}. Such positive effects are the primary reason behind the idea to use interactive robots in public-facing events, as described in Sec. 3, aiming to bring the learning environment closer to prospective students and engage them with the programme.

Peixoto et al.~\cite{peixoto2018diversity} examine approaches to support diversity in STEM and find that making use of robots in schools is a potential solution. They argue that educational robotics is well suited to support inclusion because it promotes the integration of different areas of knowledge, connects abstract and applied knowledge, facilitates systematic thinking, and creates a welcoming and playful learning environment. This aligns with the findings in~\cite{sobrany2021optimising} that suggest the creation of inclusive spaces to achieve better inclusion of Black, Asian and minority ethnic (BAME) students in higher education. Moreover, educational robotics provides a multi-modal learning experience that can cater for a diverse audience with different learning preferences and prior subject knowledge. We argue that this learning experience can be recreated in a compassionate approach~\cite{maratos2019improving} using robotics to create a welcoming environment also during public outreach events.

Following a similar line of argumentation,~\cite{plaza2020educational} present a course that uses educational robotics within an inclusive framework to strengthen the presence of underrepresented groups in STEM subjects. Their course revolves around solidifying skills like teamwork, leadership and confidence, promotion of entrepreneurship, logical thinking, curiosity, and others using robotics as the area for learning.

In a comparative study,~\cite{jackson2020soft} investigate the benefits of teaching soft robotics, a more recent development in robotics engineering, versus a traditional, rigid robotics course. While both courses promoted an interest in robotics, the soft robotics approach was able to further mitigate gender differences with an emphasis on materiality and iterative design. The authors conclude that soft robotics represents a promising and inclusive approach to teaching robotics, extending the mindset of students of what constitutes a machine, and promoting self-efficacy in tinkering. In summary, using such an approach can directly challenge stereotypes mentioned by~\cite{master2016computing} and therefore might be able to have a positive effect on student intake by raising girls’ interest in computing subjects.

Interactive robots have the potential to facilitate inclusion, for example by demonstrating culturally relevant gestures to foreign children using their limbs~\cite{de2020socially}. Similarly,~\cite{tarres2021qui} present a novel robotic challenge for children using Lego bricks aimed at breaking stereotypes and reducing barriers in children of school age. Both approaches aim to create a welcoming environment and an open and inclusive atmosphere, which facilitates inclusion amongst the learning group~\cite{sobrany2021optimising}.

In summary, the introduction of interactive robots into the classroom can facilitate a multimodal, hands-on experience that improves learning across all genders, social and ethnical backgrounds, and potential disabilities. Moreover, if used correctly, robots have the potential to close the gaps between groups by incorporating strategies that selectively raise interest or focus on addressing special needs or preferences that have been identified within the groups. In our case study, we use these insights to create an experience that aims to attract and raise the interest of visitors in a compassionate way that especially includes underrepresented groups in computer science subjects.

\begin{figure}
  \includegraphics[width=\linewidth]{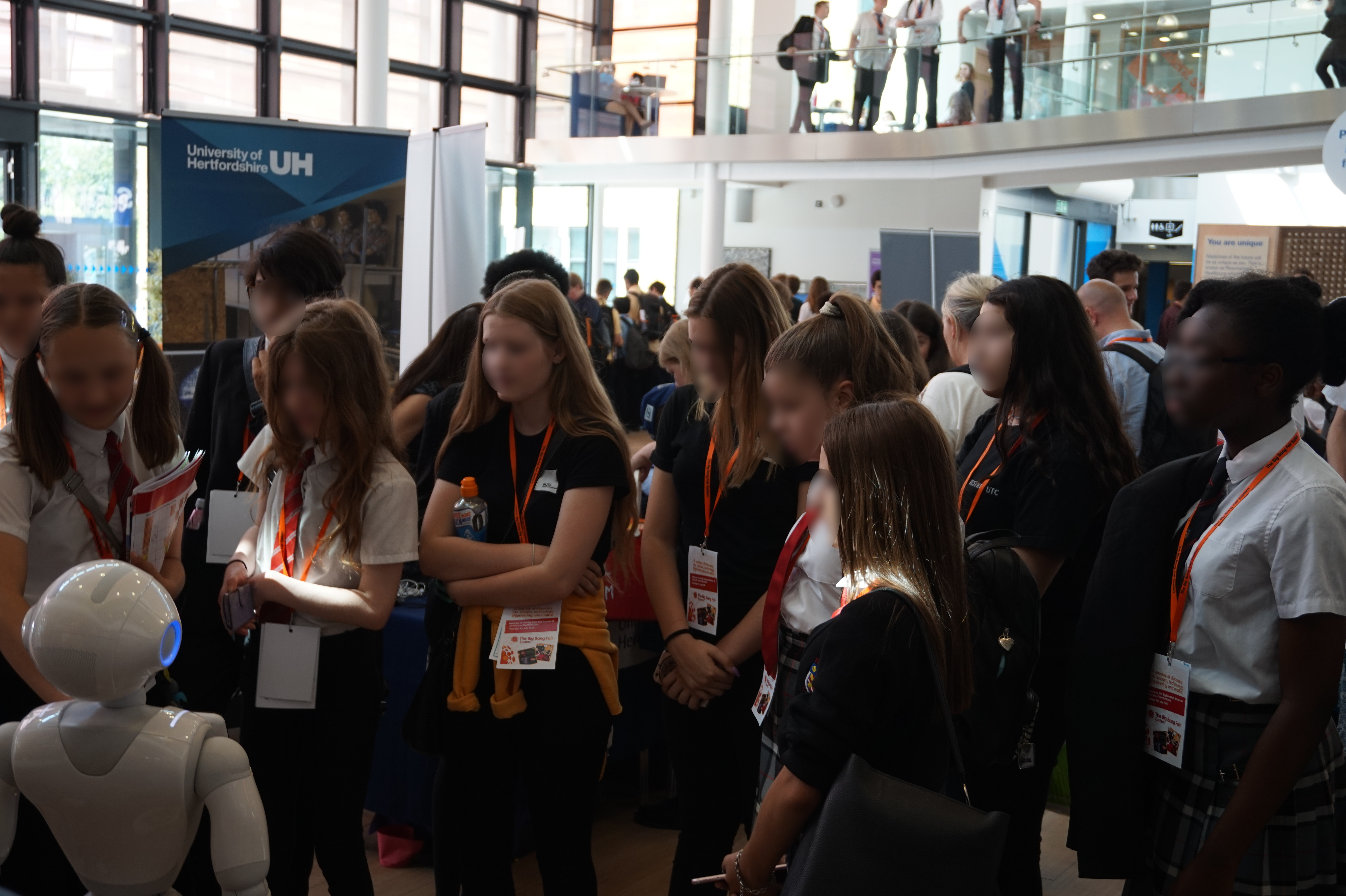}
  \caption{Group of interested people interacting with the Pepper robot at the Big Bang Fair Eastern 2018.}
  \label{fig:crowd}
\end{figure}

\section{Case Study}

In the following, we will present two case studies where a team of robotics researchers from the Adaptive Systems Research Group at the University of Hertfordshire each joined an outreach event (before the outbreak of the COVID pandemic) with the aim to attract future students to enrol in a computing subject at the university. In both events, researchers exhibited multiple robots, including the interactive robots Kaspar~\cite{Wood2021Kaspar} and Pepper~\cite{Pandey2018Pepper} for approximately three hours at a fixed booth providing the possibility to live-interact with the robots, observe their capabilities, and gain insights about their inner workings using a screen. They also introduced visitors to artificial intelligence and robotics research and explained how the robots are used in human-robot interaction experiments. General feedback for both events was incredibly positive from students, teachers, and organisers alike.

On one occasion, the booth was a part of the event ".helloworld - Hertfordshire Hackathon" organised by the British Computer Society. This event further displayed classic computing devices and other interesting, fun, and educational stalls and exhibits suitable for all ages, experience levels and interests. The event was free of charge but required prior registration and was attended by approximately 200 people. The second occasion was the Big Bang Fair Eastern 2018 with more than 2,300 visitors, aimed directly at inspiring prospective students to enrol in STEM subjects at the University of Hertfordshire. The exhibit was placed in a large hall together with colleagues from the School of Engineering and Technology, presenting a flight simulator, scanners, and a racing car. Especially in the latter event, the robots always drew enormous amounts of attention to the booths with at least one of the researchers in interaction with a visitor as shown in Figure~\ref{fig:crowd}.

In both events, one aim for the exhibitors was to put together a diverse team to allow prospective students from all backgrounds to identify with them as STEM role models, which are known to matter for underrepresented student groups at a relevant age~\cite{prieto2020stem,tyler2018engaging}. This inspired the formation of a team of researchers from diverse ethnicities, including seven national backgrounds from Europe and Asia, a fair gender ratio of three females to four males, and multiple disciplinary backgrounds while all the exhibitors were non-disabled. Moreover, the interactive robots as part of the exhibition team have been carefully selected to highlight the transdisciplinarity and inclusivity of robotics research.

\begin{figure}
  \includegraphics[angle=90, trim={0, 2cm, 3cm, 0}, clip, width=.66\linewidth]{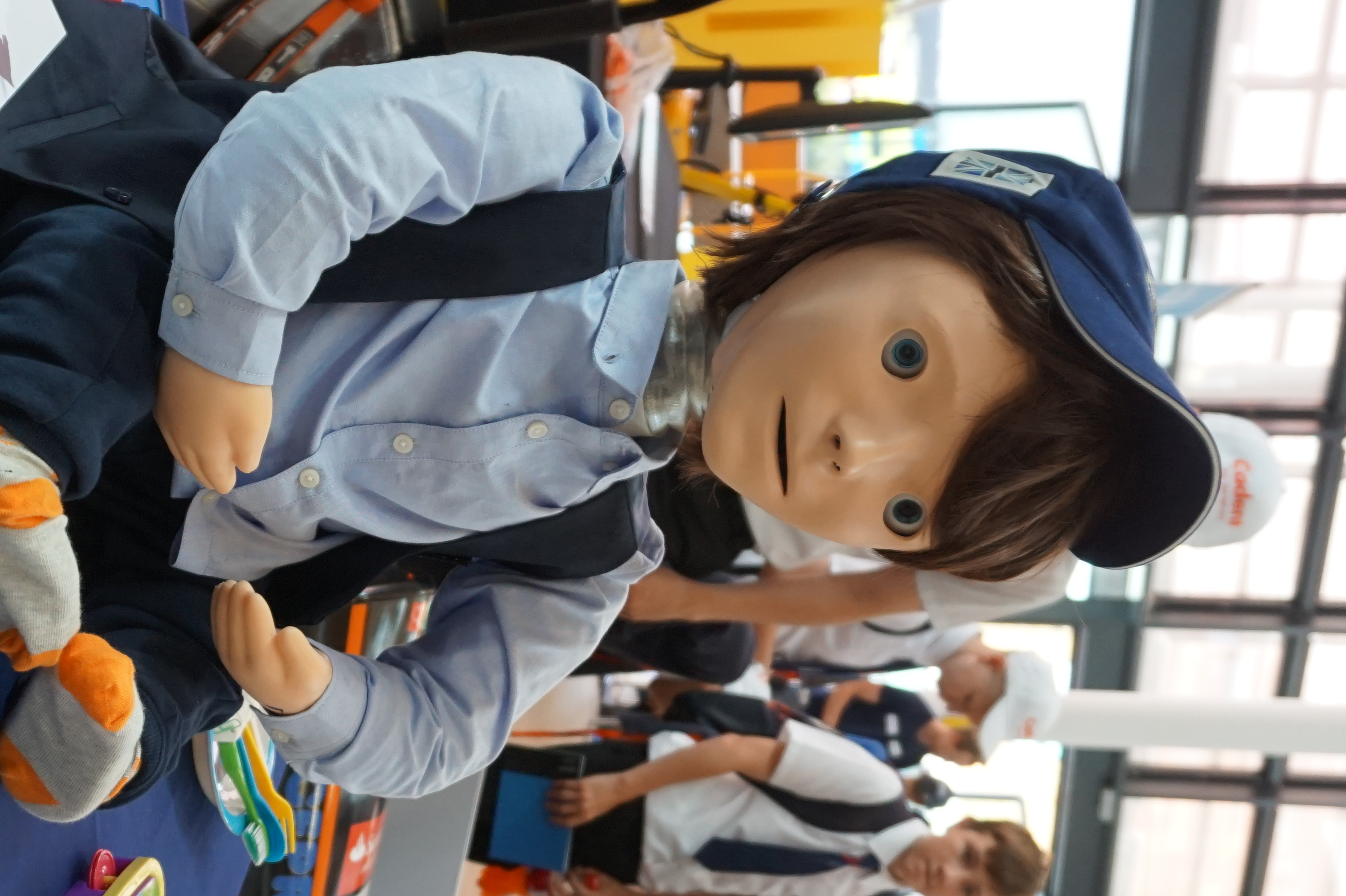}
  \caption{The Kaspar robot at the Big Bang Fair 2018.}
  \label{fig:kaspar}
\end{figure}

The Kaspar robot (Fig.~\ref{fig:kaspar}) is a state-of-the-art humanoid that has been developed for interaction with children with autism spectrum disorder and language therapy~\cite{Wood2021Kaspar}. It has been purposefully designed as an expressive tool to initially offer a more repetitive form of communication, which aims to make social interaction simpler and more comfortable for the child~\cite{dautenhahn2009kaspar}. It has been proven effective in various interactive experiments with children with special education needs to help break their social isolation by acting as a social mediator. As such, the robot can be used to interactively highlight and facilitate discussion about the different facets of assistive, social, and educational robotics. It also provides a crucial point of identification for students with special educational needs as well as for their parents or siblings.

The second robot at the exhibits, Pepper, is an industrially designed humanoid that aims to attract people into all kinds of businesses~\cite{pandey2018mass}, for example, mobile phone stores, restaurants, or shopping malls. As such, it is predestined to draw significant amounts of attention when presented at public events. Since its introduction in 2014, it has found widespread adoption in the human-robot interaction research community and has been used for investigating primarily social phenomena like trust or emotion. These research topics, like the ones described above, have the potential to connect robotics research to various other disciplines, ranging from ethology and neurobiology to industrial design or artificial intelligence. Pepper has been created to be gender-neutral~\cite{pandey2018mass}, following the Japanese phrase “docchi demo ii” meaning “either is fine”. This deliberate design approach leaves it up to the person interacting with the robot to assign a gender, with many people associating a female gender due to its voice~\cite{mcginn2019can}. The resulting characteristics and looks are important for creating an inclusive atmosphere for all genders when using the robot at a public event.

Both events facilitated short- and long-term discussions between a very diverse set of interested students and their parents on the one side and the exhibition team and the robots on the other side. Often, students spotted the robots from afar and stood around the exhibit while the robots were, for example, doing a dance routine, moving their artificial limbs, or telling a story. Using such performances, the two robots in both events were able to facilitate group situations, much like a performer on stage, between people waiting and kept them entertained whilst some of the presenters were in conversation with other visitors. Interested proactive people would then often go into a more focused situation with a team member and the robot with the others being able to continue to observe. Sometimes, students seemed interested but needed encouragement or initiative from their parents to engage in conversation and other times, students were approached directly by one of the team. There have been cases where parents were fascinated by the progress of autism research and how assistive robotics can help to address the specific needs of individuals as opposed to operating in large industries. Across genders and ethnicities, visitors have been surprised that robotics research encompasses such a vast range of topics and is not constrained to mechanical engineering.

From the team’s perspective, both events were incredibly successful in increasing the visibility of the research and teaching content to a lay audience. At the same time, curious minds that are not exposed to the topic on a daily basis, no matter if accompanying parents or prospective students, can provide valuable novel perspectives, initiate new research ideas, or seriously question current approaches. The computer science department is also happy with the received attention frequently asking the team to be involved in similar activities.

\section{Discussion}
When looking back at the presented case study, the conduction of a successful and inclusive outreach activity to promote robotics for a diverse group of university starters seems to benefit greatly from the combination of a diverse team with two entirely different robots that have unique and wide applications. One of the primary approaches here could be to reduce stereotypes about robotics as a field. To be successful in reducing such stereotypes, it is important to extend the mindset of the people who might be applying them. In this case, by relating the core topic of robotics to applications and problems in other fields of research, to real-world applications, or to others’ personal lives. Besides extending others’ mindsets about stereotypes in robotics, there is also the possibility to associate the field with people who defy the stereotype by appearing as a diverse set of successful researchers. A diverse team of presenters, at the same time, underlines the validity of a career choice in robotics for interested students from divergent ethnic backgrounds or underrepresented genders. Highlighting the significant role of robotics in autism research also constitutes an inclusive approach for potential students with disabilities and their families and helps them relate to the field of robotics.

\subsection{Implications}
The wide range of student-facing activities that the University of Hertfordshire engages in include, amongst others, open days, fairs, and exhibitions. The university states that its outreach activities "[aim] to inspire and to take away any confusion” underlining such activities are vital in shaping their student intake by informing and attracting future students~\cite{UH2022ELO}. This is in line with~\cite{heaslip2020situating} who find, in a systematic review, that outreach can be successful in widening participation in higher education.

The presented case study based on two such outreach activities at open days yields the following action recommendations for other higher education institutions to promote computing courses:
\begin{itemize}
 \item Be visible: Using interactive robotic technology can draw a significant amount of attention to your exhibit and facilitate interacting with the visitors.
 \item Be relatable: Representing the diverse reality of the university and research team plus the readiness to discuss a wide range of topics can increase how much the visitors can identify with the field.
\end{itemize}
In summary, the composition of the team and the interplay between the team, robots, and visitors plays a vital role in successfully engaging the visitors.

\subsection{Limitations}
The presented case study has several limitations that have the potential to be addressed in follow-up research. Most importantly, the study gives no direct estimate on the effectiveness of using a robot at public events versus not using one. A comparative quantitative study between two independent events or booths, one with and one without using interactive robots, would be required to directly measure such an effect. However, the presented qualitative case study provides insights into the apparent benefits of having interactive robots at the booth.

A second shortcoming of the presented study is that it is not exhaustive of underrepresented groups in STEM subjects. It, for example, gives some evidence about being inclusive towards people with autism but lacks insights into whether that is translatable to other disabilities. Moreover, the current study does not discuss whether having a disabled person as a part of the team would have a positive effect on inclusivity, which we assume that it should have a positive effect.

The study did not take into account the relation of money and time spent on the activity versus the potential income, for example in tuition fees. Former involves a significant number of staff spending time preparing for and during the exhibit. Given added programming and testing, preparation time is most likely higher compared to a booth that does not rely on robotic technology. However, from our experience, that time can be greatly reduced when planning for presenting at subsequent events and a routine has been found. The robots themselves also introduce a non-insignificant cost, consisting of purchasing and ongoing support. This might pose a problem for computer science departments that do not have any relation to robotics but in other departments could be covered by existing infrastructure. Some interactive robots, however, are mass-produced and can be purchased at a comparatively low price.

\section{Conclusion}
In this paper, we have discussed how two outreach events at the University of Hertfordshire were successful in compassionately engaging potential students. We have identified two recommendations to raise interest across genders, social and ethnic backgrounds, and potential disabilities: Interactive robots proved well-suited as inclusive tools in being able to engage students in different sub-disciplines of computer science attracting people using their interactive capabilities. In combination with a diverse team of exhibitors, an inclusive atmosphere was successfully established. The events facilitated the reduction of stereotypes and increased the curiosity of potential students, which is key in increasing interest among underrepresented groups in computer science subjects. The discussion further included limitations of the presented study in missing evidence about the measurable success of the approach, which we encourage to address systematically in follow-up research.

\bibliographystyle{ACM-Reference-Format}
\bibliography{bib}

\end{document}